\documentclass[runningheads]{llncs}
 
\usepackage{eccv}
\usepackage{eccvabbrv}
\usepackage{graphicx}
\usepackage{booktabs}
\usepackage{multirow}
\usepackage{xcolor}
\usepackage{colortbl}
\usepackage{amsmath}

\usepackage[accsupp]{axessibility}  
\usepackage{hyperref}
\usepackage{orcidlink}
\usepackage{bm}
\usepackage{pifont}

\definecolor{eccvblue}{rgb}{0.12,0.49,0.85}
\definecolor{myred}{RGB}{180,40,40}
\definecolor{myblue}{RGB}{35,90,160}
\definecolor{mygray}{RGB}{90,90,90}
\definecolor{darkgreen}{RGB}{0,128,0}

\begin{document}

\title{Mitigating Positional Leakage in 3D Masked Autoencoders for Robust Representation Learning}

\titlerunning{MPL-MAE}

\newcommand{\samethanks}[1][\value{footnote}]{\footnotemark[#1]}
\author{%
  Xu Yan\inst{1,2}\thanks{Equal contribution.} \and
  Huiqun Wang\inst{1,2}\samethanks \and
  Chen Wang\inst{3}\textsuperscript{$\dagger$} \and
  Lei Ren\inst{4} \and
  Di Huang\inst{1,2}\textsuperscript{$\dagger$}
}
\authorrunning{Y. Xu et al.}

\institute{
  SKLCCSE, Beihang University, Beijing, China \and
  SCSE, Beihang University, Beijing, China \and
  NERCBDS, EIRI, Tsinghua University, Beijing, China \and
  SASEE, Beihang University, Beijing, China \\[6pt]
  \email{\{yanx57, hqwangscse, renlei, dhuang\}@buaa.edu.cn, wang\_chen@tsinghua.edu.cn}
}

\maketitle
\let\thefootnote\relax\footnotetext{$\dagger$~Corresponding author.}

\begin{abstract}
Masked autoencoding has emerged as a prominent paradigm for self-supervised learning on 3D point clouds, achieving competitive performance across downstream tasks. Unlike its 2D counterpart, 3D masked autoencoding directly reconstructs spatial coordinates, making it inherently susceptible to positional leakage. In this work, we identify that the decoder in existing 3D MAE frameworks tends to over-rely on positional information, which weakens semantic representation learning and leads to suboptimal feature quality. To address this issue, we propose MPL-MAE, a masked point learning framework that mitigates positional over-reliance while enhancing the utilization of encoder features. Specifically, we introduce a recalibrated positional embedding module that suppresses metric-dominant coordinate signals while preserving geometric topology, together with a gated positional interface module that dynamically regulates positional injection during reconstruction. These designs promote a more balanced interaction between spatial priors and semantic features, yielding robust and informative representations. Extensive experiments across downstream tasks demonstrate that MPL-MAE consistently achieves competitive performance, validating its effectiveness. Code is available at \url{https://github.com/yanx57/MPL-MAE}.
\end{abstract}

\section{Introduction}
\label{sec:intro}
Self-supervised learning for 3D point clouds has attracted increasing attention, as it mitigates the high costs of large-scale 3D data acquisition and manual annotation that constrain supervised approaches. It has been widely adopted in applications such as autonomous driving~\cite{sautier2022image,luo2021self,mersch2022self}, robotics~\cite{yu2024robotic,peng2022self}, and augmented reality~\cite{wu2023masked,wu2020pointpwc}.

Among various self-supervised paradigms, masked autoencoders (MAE), originally proposed for 2D visual representation learning, have been extensively adapted to 3D point cloud representation learning due to their simplicity and effectiveness. Existing 3D MAE methods~\cite{pointbert, pointmae, point-m2ae, pcp-mae, pointfemae} mask a substantial portion of the input point cloud and train the model to reconstruct the missing content from limited visible context, demonstrating strong empirical performance.


However, despite their success, a fundamental discrepancy between 2D images and 3D point clouds poses challenges to the direct adoption of MAE. Unlike 2D images, which are structured as ordered pixel grids, point clouds are unordered sets of discrete coordinates. In MAE, reconstruction supervision typically combines positional embeddings of masked regions with encoder representations to predict missing content. When applied to point clouds, the coordinates of masked regions are directly tied to the reconstruction targets and may provide deterministic geometric cues about the missing content. This can introduce information leakage and weaken the effectiveness of self-supervised learning.

\begin{figure}[t]
    \centering
    \begin{subfigure}[t]{0.37\linewidth}
    \vspace{0pt}
        \centering
\includegraphics[width=\linewidth]{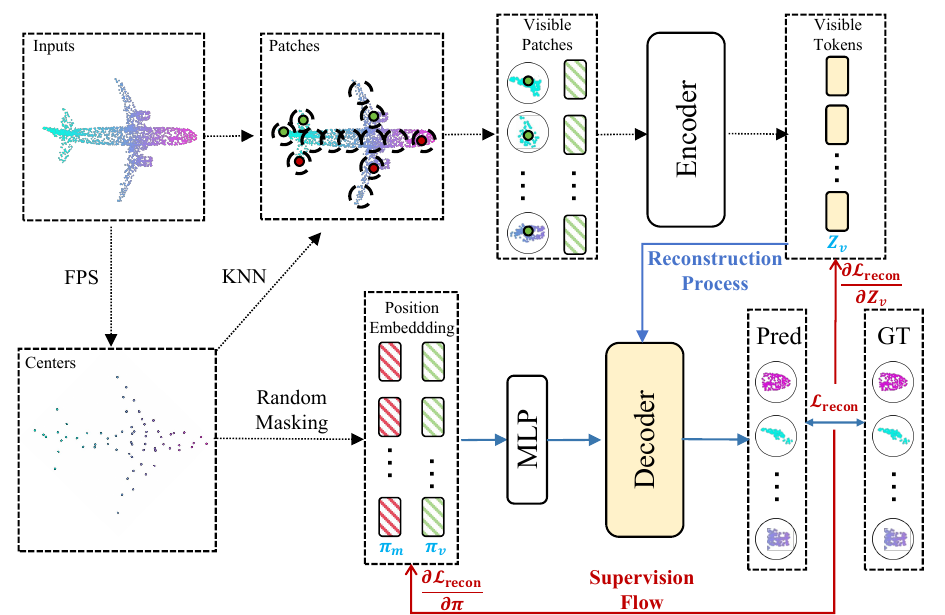}
        \caption{}
    \end{subfigure}
    \hspace{-0.02\linewidth}
    \begin{subfigure}[t]{0.62\linewidth}
    \vspace{0pt}
        \centering
        \includegraphics[width=\linewidth]
         {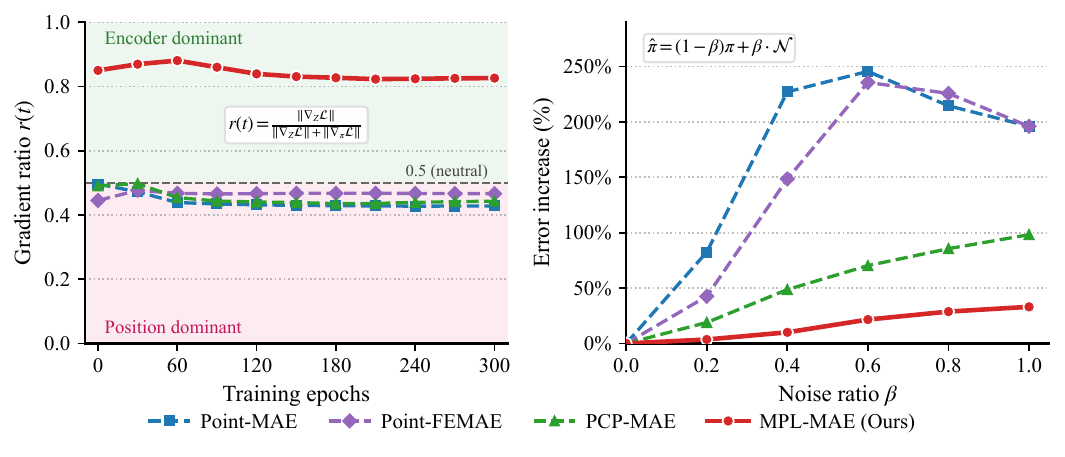}
        \caption{}
    \end{subfigure}
\caption{
(a) Pipeline of MAE for 3D point cloud representation learning. 
(b) Empirical analysis of representative 3D MAE methods. 
Left: gradient ratio between positional embeddings ($\pi$) and encoder outputs ($Z$) during training. 
Right: sensitivity analysis of reconstruction error under perturbations applied to positional embeddings.
}
    \label{fig:fig1}
\end{figure}

To further investigate this issue, we conduct a detailed analysis of both the pre-training dynamics and the resulting pre-trained encoder. As illustrated in Fig.~\ref{fig:fig1}, under the reconstruction supervision adopted by mainstream 3D MAE frameworks, the gradients associated with positional embeddings consistently exhibit larger magnitudes than those of the encoder outputs throughout optimization. This imbalance suggests an optimization bias: the model tends to over-rely on positional embeddings while under-utilizing the semantic features learned by the backbone. Although positional information is necessary for masked reconstruction, excessive metric-coordinate cues may enable the decoder to bypass encoder-derived semantics, making reconstruction highly dependent on positional embeddings rather than encoder outputs. As a result, when positional embeddings are perturbed with noise, the reconstruction quality collapses.

Concurrent improvements, including modified encoder processing strategies~\cite{pointmae}, auxiliary supervision~\cite{pcp-mae}, and coordinate decoupling~\cite{chen2023pointgpt}, have improved empirical performance but still exhibit similar behavior, suggesting that they remain susceptible to the same positional shortcut. These observations indicate that the model may converge to a shortcut solution that prioritizes spatial cues provided by positional embeddings over learning meaningful semantic representations.

To address this issue, we reform the utilization of positional embeddings in the decoding stage of MAE to enforce more semantically grounded reconstruction. In 3D MAE, positional embeddings are intended to provide structural cues that guide spatial reasoning in the decoder, rather than directly encoding explicit coordinate information that enables trivial reconstruction. Achieving this objective requires two key principles: (1) positional embeddings should preserve the intrinsic geometric topology of the point cloud, and (2) they should avoid encoding explicit absolute coordinates that dominate the optimization process.


Motivated by these insights, we propose a Recalibrated Positional Embedding (RPE) module that preserves topological structure while suppressing coordinate-specific signals. We further introduce a Gated Positional Interface (GPI) module in the decoder to dynamically regulate the contribution of positional embeddings during reconstruction. By adaptively balancing geometric cues and semantic features, GPI prevents the decoder from over-relying on positional shortcuts. Integrating these components, we establish MPL-MAE, a novel MAE framework for 3D representation learning that mitigates positional over-reliance in conventional MAE architectures and promotes stronger semantic representation learning. As a result, MPL-MAE achieves improved downstream performance and enhanced robustness under noisy or perturbed conditions.

Our main contributions are summarized as follows:

\begin{itemize}
\item We propose a RPE module to suppress coordinate-specific signals while preserving geometric topology, and a GPI module to dynamically regulate positional injection during training.
\item We integrate these components into MPL-MAE, a novel masked autoencoder framework that mitigates positional shortcut learning and promotes balanced semantic representation learning.
\item We conduct extensive experiments across multiple benchmarks, demonstrating competitive or superior downstream performance and improved robustness compared with existing 3D MAE variants.
\end{itemize}

\section{Related Work}
\subsection{Contrastive-based Self-Supervised Learning}

Contrastive-based self-supervised methods~\cite{pointcontrast, csc, chhipa2022depthcontrastselfsupervisedpretraining, huang2021spatio, afham2022crosspoint} learn 3D representations by pulling together augmented views of the same instance while pushing apart different samples. PointContrast~\cite{pointcontrast} leverages spatial correspondences between overlapping partial scans to learn viewpoint-invariant local features. CSC~\cite{csc} extends this idea by enforcing both point-level and partition-level consistency to capture local–global contextual alignment. DepthContrast~\cite{chhipa2022depthcontrastselfsupervisedpretraining} proposes a format-agnostic framework that aligns global features across point clouds and depth maps, but inherits the dependence of contrastive learning on large batch sizes and carefully constructed negatives. STRL~\cite{huang2021spatio} alleviates this issue by adopting a BYOL-style~\cite{byol} self-distillation mechanism, improving efficiency without explicit negative pairs, although its representations remain largely geometry-driven.
CrossPoint~\cite{afham2022crosspoint} introduces cross-modal contrast by aligning 3D point clouds with 2D image projections, enriching 3D features with semantic priors. Subsequent works~\cite{ulip, ulip2, uni3d, mrd} further explore multimodal alignment strategies to learn unified 3D representations.

Despite their success, contrastive approaches heavily depend on augmentation design and alignment strategies, primarily enforcing instance-level discrimination or cross-modal consistency. They do not explicitly model geometric reconstruction, which may limit their ability to fully exploit intrinsic structural cues within point clouds. This motivates reconstruction-based paradigms such as masked autoencoders.

\subsection{Reconstruction-based Self-Supervised Learning}

Reconstruction-based methods~\cite{pointbert, pointmae, point-m2ae, pointfemae, pcp-mae} learn 3D representations by recovering masked inputs within an encoder–decoder framework.
Point-BERT~\cite{pointbert} introduces a discrete variational autoencoder to tokenize point patches into visual words and formulates masked token prediction as a classification task. However, its reliance on a pre-trained tokenizer increases computational complexity and introduces dependency on codebook quality. Point-MAE~\cite{pointmae} simplifies this pipeline by directly reconstructing raw point coordinates, demonstrating that an end-to-end regression objective can yield strong representations.
Subsequent works enhance architectural design or supervisory signals. Point-M2AE~\cite{point-m2ae} adopts a hierarchical architecture to capture multi-scale geometry. Point-FEMAE~\cite{pointfemae} refines masking strategies with dual-branch global and local reconstruction to strengthen semantic extraction. PCP-MAE~\cite{pcp-mae} augments supervision by predicting geometric centers alongside coordinates to enhance positional awareness. PQAE \cite{zhang2025diversechallengingpretrainingpoint} explores a cross-reconstruction paradigm by generating two decoupled point cloud views and reconstructing one from the other. Recent studies further extend masked reconstruction to multimodal settings~\cite{dong2023autoencoderscrossmodalteacherspretrained, guo2023joint, zhang2023learning, chen2023pimae, liu2024inter} and broader 3D tasks~\cite{liu2026dcmae, min2023occupancy, li2022semmae}.

Despite these advances, existing reconstruction-based approaches primarily focus on architectural refinements or enhanced supervision while overlooking the interaction between positional embeddings and semantic features, which can lead to positional shortcut learning. This limitation motivates our investigation into positional leakage and the development of mechanisms that rebalance geometric cues and semantic representation learning.

\section{Method}

\subsection{Preliminary}

Given an input point cloud $X \in \mathbb{R}^{n \times 3}$, 3D MAE first partitions it into a set of local patches. Specifically, a subset of centroids is sampled using Farthest Point Sampling (FPS). For each centroid, its neighboring points are grouped via $K$-nearest neighbors (KNN), producing $G$ local patches.

A predefined masking ratio is then applied to these patches. The visible patches, denoted as $P_v$, are fed into a tokenizer and subsequently passed to the encoder together with their corresponding positional embeddings. The encoder produces latent representations $Z_v$ for the visible patches. These representations, along with learnable masked tokens and positional embeddings associated with the masked patches, are then fed into the decoder, which reconstructs the masked point sets. The reconstruction objective is optimized using the $\ell_2$ Chamfer Distance (CD)~\cite{fan2017point}, defined as:
 \begin{equation}
 \mathcal{L}_{\mathrm{Recon}} =
 \frac{1}{|P_{\mathrm{pred}}|} \sum_{a \in P_{\mathrm{pred}}} \min_{b \in X_m} |a-b|_2^2
 +
 \frac{1}{|X_m|} \sum_{b \in X_m} \min_{a \in P_{\mathrm{pred}}} |a-b|_2^2,
 \end{equation}
 where $P_{\mathrm{pred}}$ denotes the reconstructed points and $X_m$ denotes the ground-truth masked points.

This paradigm has been widely adopted in recent works~\cite{pointmae, point-m2ae, pcp-mae, pointfemae}. However, during reconstruction, the positional embeddings of masked patches may carry explicit spatial cues about the target locations, potentially leading to positional leakage. To characterize this effect, we consider the decoder input $H = [Z; \pi ]$,
where $\pi$ denote the positional embeddings for patches and $Z$ represents the semantic tokens. Under the reconstruction loss $\mathcal{L}_{\mathrm{Recon}}$, we monitor the relative gradient magnitudes in Fig.~\ref{fig:fig1} (b)

\begin{equation}
r(t) =
\frac{\left| \nabla_{Z} \mathcal{L}_{\mathrm{Recon}}(t) \right|_2}
{\left| \nabla_{Z} \mathcal{L}_{\mathrm{Recon}}(t) \right|_2
+
\left| \nabla_{\pi} \mathcal{L}_{\mathrm{Recon}}(t) \right|_2 }.
\end{equation}

Compared with the dynamically evolving and task-dependent representations in $Z$, positional embeddings provide more stable and directly exploitable spatial signals, as they encode explicit coordinate information that remains consistent throughout training. As a result, the gradient magnitude with respect to $\pi$ tends to dominate, indicating that the decoder increasingly relies on positional embeddings rather than semantic features from the encoder. This imbalance encourages the model to minimize the reconstruction loss through positional shortcuts, leading to positional leakage and suboptimal semantic representation learning during training. Consequently, the encoder is insufficiently incentivized to learn discriminative geometric features, which limits downstream generalization performance.

\subsection{Recalibrated Position Embedding}


In point cloud representations, positional embeddings typically contain two types of information: topological information and metric information. Topological information characterizes intrinsic neighborhood relationships and geometric connectivity among local patches, reflecting the structural organization of the point cloud. In contrast, metric information encodes absolute spatial coordinates and precise Euclidean distances in the ambient space. While topological information provides meaningful structural priors that facilitate geometric reasoning, metric information exposes explicit coordinate cues that can be directly exploited by the decoder for trivial reconstruction. To address this issue, we propose a Recalibrated Positional Embedding (RPE) module that preserves topological structure while suppressing metric information in 3D positional encoding.

\noindent \textbf{Order-Isomorphism Encoding.}
To mitigate coordinate leakage, we avoid directly using the continuous coordinates of each patch center. Instead, we replace them with quantized ranking indices along the $x$, $y$, and $z$ axes. Formally, let $\{c_i\}_{i=1}^{G}$ denote the set of patch centers, where $c_i = (x_i, y_i, z_i)$. For each axis $d \in \{x, y, z\}$, we define a ranking function
\begin{equation}
    r_i^{(d)} = \mathrm{Rank}\big(c_i^{(d)} \;\big|\; \{c_j^{(d)}\}_{j=1}^{G}\big),
\end{equation}

\noindent where $r_i^{(d)} \in \{1, \dots, G\}$ represents the ordinal position of the $i$-th patch center along axis $d$. The patch center is thus represented by a discrete triplet
$\mathbf{r}_i = (r_i^{(x)}, r_i^{(y)}, r_i^{(z)}).$
This order-isomorphic transformation preserves the relative ordering of patch centers along each axis while discarding absolute metric information. Specifically, for any two patches $i$ and $j$, the relation
\begin{equation}
x_i < x_j \quad \Leftrightarrow \quad r_i^{(x)} < r_j^{(x)}
\end{equation}
holds, ensuring monotonic invariance under any strictly increasing transformation of the coordinates. Consequently, the embedding depends solely on the ordering structure of the point cloud rather than its absolute spatial scale. Since precise metric distances are no longer explicitly encoded, the decoder cannot exploit coordinate shortcuts for trivial reconstruction.

To embed the discrete rank indices into a continuous space, we apply sinusoidal positional encoding to each axis:
\begin{equation}
\phi(r_i^{(d)}, 2k) = \sin\!\left(\frac{r_i^{(d)}}{\tau^{2k/D}}\right), 
\quad
\phi(r_i^{(d)}, 2k+1) = \cos\!\left(\frac{r_i^{(d)}}{\tau^{2k/D}}\right),
\end{equation}
where $k$ indexes the embedding dimension, $D$ denotes the embedding size per axis, and $\tau$ is a temperature hyperparameter controlling frequency scaling. The final encoded representation for patch $i$ is obtained by concatenating the embeddings from all three axes:
\begin{equation}
\tilde{\mathbf{e}}_i = \mathrm{Concat}\big(\phi(r_i^{(x)}), \phi(r_i^{(y)}), \phi(r_i^{(z)})\big).
\end{equation}
Finally, a lightweight residual multi-layer perceptron $f$ projects the encoded representation into the latent embedding space:
$
\tilde{\mathbf{\pi}_i} = f(\tilde{\mathbf{e}}_i),
$
yielding the recalibrated positional embeddings used by the decoder. This design preserves coarse geometric topology through ordinal structure while suppressing coordinate-specific signals that would otherwise dominate the optimization process.

\noindent \textbf{Topology Preservation Regularization.}
 Beyond order-isomorphism encoding, we introduce a unified regularization objective to explicitly preserve geometric topology. The goal is to ensure that the recalibrated positional embeddings retain structural adjacency relationships among patches.

Let $\pi_i$ denote the original positional embedding derived from the $i$-th patch center, and let $\tilde{\pi}_i$ denote its recalibrated counterpart produced by RPE. To preserve structural consistency, we maintain local neighborhood relations. Let $\mathcal{N}_k(\pi_i)$ denote the set of $k$-nearest neighbors of patch $i$ in the original positional embedding space. We enforce local topological consistency through
\begin{equation}
\mathcal{L}_{\mathrm{topo}} =
\frac{1}{|V|}
\sum_{i \in V}
\sum_{j \in \mathcal{N}_k(\pi_i)}
\left(
d(\tilde{\pi}_i, \tilde{\pi}_j)
-
d(\pi_i, \pi_j)
\right)^2,
\end{equation}
where $V$ denotes the set of all patch indices and $d(\cdot,\cdot)$ is a distance metric such as Euclidean or cosine distance.

This regularization term preserves relative adjacency relationships among patches and provides a structural scaffold for spatial reasoning without enforcing absolute metric fidelity. By constraining the recalibrated embeddings to respect local topology while discarding explicit coordinate magnitudes, the decoder is prevented from minimizing the reconstruction objective through direct metric cues. As a result, the model is encouraged to rely more heavily on encoder-derived semantic representations, promoting more balanced and semantically grounded feature learning.

\subsection{Gated Positional Interface}
During the reconstruction phase, positional embeddings of masked patches are injected into every self-attention layer of the decoder by default. However, such intensive and unconditional incorporation of positional information may reinforce positional dominance, causing the decoder to over-rely on spatial cues and further aggravating shortcut learning. To address this issue, we redesign the interface through which positional embeddings interact with decoder features.

\noindent \textbf{Dynamic Gated Integration.} Instead of the unconditional additive injection used in standard Transformers, we introduce a dynamic gated function to regulate the interaction between latent features and positional cues. In conventional designs, positional embeddings are directly added to feature representations at each layer. In contrast, we redefine the input $U_l$ to the attention mechanism of the $l$-th decoder layer as
 \begin{equation}
 U_l = H_{l-1} + \mathcal{G}_l \odot \tilde{\pi},
 \end{equation}
 where $H_{l-1}$ denotes the feature state from the previous layer, with $H_0 = Z$, and $\mathcal{G}_l$ is a lightweight gating network that produces element-wise modulation coefficients.

This formulation casts positional injection as a differentiable selection mechanism, encouraging the decoder to adaptively determine whether to utilize positional guidance or rely on latent semantic features. Rather than always exposing the decoder to positional signals, the gate dynamically controls both the presence and the strength of positional information at each layer and token location.

For each decoder layer, the gating vector $\mathcal{G}_l$ is sampled from a categorical distribution over two states, namely suppression and injection. Specifically, we compute
\begin{equation}
 \mathcal{G}_l = \mathrm{Gumbel\text{-}Softmax}\big(\Phi_\theta(\tilde{\pi}), \tau),
 \end{equation}
where $\Phi_\theta$ denotes a lightweight mapping that produces categorical logits, and $\tau$ is a temperature parameter controlling the sharpness of the distribution. During the forward pass, the gate yields a near one-hot vector via the straight-through estimator, thereby stochastically restricting the bandwidth of the positional signal at each token location.

By enforcing sparse and selective positional injection, the decoder is prevented from consistently exploiting positional shortcuts. This mechanism promotes a more balanced interaction between geometric priors and encoder-derived semantic representations, thereby mitigating positional leakage and encouraging semantically grounded reconstruction.

\noindent \textbf{Leakage Regularization.} In addition to dynamic gated integration, we further suppress residual metric leakage to ensure that positional embeddings remain insufficient for direct coordinate reconstruction.

To prevent positional shortcuts, we impose a metric leakage constraint that limits the reconstructability of masked coordinates from positional embeddings alone. For a probe function family $\mathcal{H}$, we require
\begin{equation}
\min_{h \in \mathcal{H}}
\mathbb{E}
\left[
\mathrm{CD}\big(X_m, h(\tilde{\pi})\big)
\right]
\ge \epsilon,
\end{equation}
where $X_m$ denotes the ground-truth coordinates of masked patches, $\mathrm{CD}(\cdot,\cdot)$ represents the Chamfer Distance and $\epsilon > 0$ is a predefined lower bound. This condition ensures that the recalibrated positional embeddings are not sufficiently informative for direct geometric reconstruction.

Directly enforcing this inequality is intractable. Therefore, we introduce a tractable surrogate objective $\mathcal{L}_{\mathrm{leak}}$ by directly using the decoder to predict masked coordinates from positional embeddings alone. Specifically, we feed $\tilde{\pi}$ into the decoder while masking out encoder features, and compute
\begin{equation}
\mathcal{L}_{\mathrm{leak}} =
-\mathrm{CD}\big(X_m, \mathrm{Dec}(\tilde{\pi})\big),
\end{equation}
where $\mathrm{Dec}(\cdot)$ denotes the reconstruction decoder.

During optimization, the decoder parameters are trained to minimize the reconstruction loss, while the positional encoding module is optimized adversarially to maximize the reconstruction error under positional-only input. This min–max formulation approximates the lower-bound constraint and explicitly reduces the coordinate predictability of recalibrated positional embeddings.

By limiting the amount of metric information recoverable from positional cues, the decoder is further discouraged from exploiting positional shortcuts. Consequently, the reconstruction process is compelled to depend more heavily on encoder-derived semantic representations, promoting balanced optimization and stronger geometric feature learning.

\label{sec:framework}
\begin{figure}[!tb]
	\includegraphics[width = 1.0\linewidth]{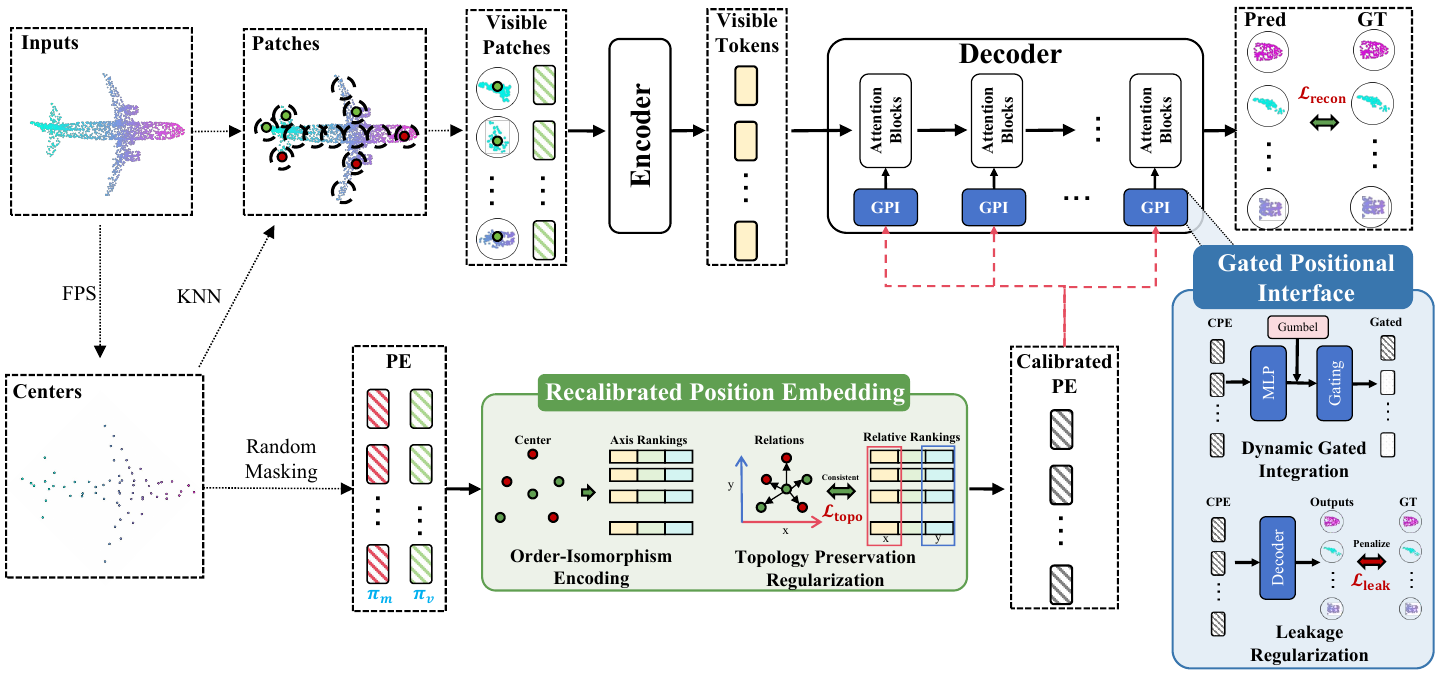}
\caption{
Pipeline of MPL-MAE. The RPE module generates recalibrated positional embeddings for the decoder. A GPI module is inserted before each Transformer layer in the decoder to regulate positional injection, promoting balanced optimization and robust geometric feature learning.
}
\label{fig:framework}
\end{figure} 

\subsection{Framework}

By integrating the Recalibrated Positional Embedding (RPE) module and the Gated Positional Interface (GPI), we present MPL-MAE, a novel masked autoencoder paradigm for 3D point clouds that mitigates positional leakage, as illustrated in Fig.~\ref{fig:framework}.

During the encoding phase, only visible patches are processed by the tokenizer and encoder to produce latent semantic representations $Z_v$. No masked positional embeddings are exposed to the encoder, ensuring that semantic feature extraction is decoupled from masked spatial cues. During the decoding phase, the decoder takes as input the visible embeddings $Z_v$, learnable masked tokens, and the recalibrated positional embeddings generated by RPE. The GPI mechanism is incorporated into each self-attention layer of the decoder to dynamically regulate the injection of positional information. This design prevents unconditional positional dominance and encourages adaptive interaction between geometric priors and encoder-derived semantic features.

The entire MPL-MAE framework is trained under a multi-task objective that simultaneously enforces reconstruction fidelity, topological consistency, and metric non-sufficiency. The overall loss function $\mathcal{L}_{\mathrm{total}}$ is formulated as
 \begin{equation}
 \mathcal{L}_{\mathrm{total}}
 =
 \mathcal{L}_{\mathrm{recon}}
 + \alpha \mathcal{L}_{\mathrm{topo}}
 + \beta \mathcal{L}_{\mathrm{leak}},
 \end{equation}
 where $\mathcal{L}_{\mathrm{recon}}$ denotes the standard Chamfer Distance (CD) or $\ell_2$ reconstruction loss used in Point-MAE for coordinate prediction of masked patches $X_m$, $\mathcal{L}_{\mathrm{topo}}$ preserves local structural adjacency in the recalibrated positional embeddings, and $\mathcal{L}_{\mathrm{leak}}$ suppresses residual metric leakage by reducing coordinate predictability from positional cues alone. The coefficients $\alpha$ and $\beta$ balance the contributions of topology preservation and leakage regularization.

By jointly enforcing these objectives, MPL-MAE mitigates positional shortcut learning while preserving meaningful geometric structure. This leads to more balanced optimization, stronger semantic representation learning, and improved downstream generalization performance.

\section{Experiments}

\subsection{Implementation Details}

\noindent \textbf{Datasets.}
To comprehensively evaluate MPL-MAE, we conduct experiments on two representative 3D perception benchmarks: ModelNet40~\cite{ModelNet40} and ScanObjectNN~\cite{scanobjectNN}. ModelNet40 contains 12,311 clean CAD models from 40 categories and serves as a standard benchmark for object-level representation learning, while ScanObjectNN comprises around 15,000 real-world scanned objects from 15 categories with occlusion, clutter, and sensor noise, providing a more challenging evaluation of generalization.

\noindent \textbf{Settings.}
Following prior works~\cite{pointmae, pcp-mae, pointfemae}, we pre-train MPL-MAE on ShapeNet~\cite{chang2015shapenetinformationrich3dmodel} with a 12-block Transformer encoder and a 4-block Transformer decoder, both using a hidden dimension of 384 and 6 attention heads. Each input point cloud is downsampled to 1024 points and divided into 64 local patches via FPS and $k$-NN, with $k=32$ points per patch. Following common protocols~\cite{pcp-mae, qi2023contrastreconstructcontrastive3d, zhang2025diversechallengingpretrainingpoint}, we apply scaling, translation, and rotation for data augmentation. The model is trained for 300 epochs using AdamW~\cite{loshchilov2017sgdrstochasticgradientdescent} with an initial learning rate of $5 \times 10^{-4}$, weight decay of 0.05, cosine decay~\cite{loshchilov2017decoupled}, and a 10-epoch linear warmup. We set $\alpha=0.05$ and $\beta=0.1$. Training is conducted on a single NVIDIA GeForce RTX 3090 GPU and takes approximately 7.5 hours.



\begin{table}[!tb]
    \centering
    \caption{Classification accuracy (\%) on ScanObjectNN~\cite{scanobjectNN} and ModelNet40~\cite{ModelNet40}. \#P~(M) denotes the number of parameters (in millions). The best results are highlighted in \textbf{bold}, and the second-best results are highlighted with \underline{underlining}. For fair comparison, we report the performance of all models from the final epoch of pre-training. The dagger($^\dagger$) denotes the baseline results which aligns augmentation with us and PCP-MAE.}
    \label{tab:one}
    
    \resizebox{\textwidth}{!}{
        \begin{tabular}{l c cccc ccc} 
        \toprule
        \multirow{2}{*}{Methods} & \multirow{2}{*}{\#P (M)} & \multicolumn{4}{c}{ScanObjectNN} & \multicolumn{3}{c}{ModelNet40} \\
        \cmidrule(lr){3-6} \cmidrule(lr){7-9}
        & & Input & OBJ-BG & OBJ-ONLY & PB-T50 & Input & w/o vote & w/ vote \\
        \midrule
        \multicolumn{9}{c}{\textit{Supervised Learning Only}} \\ 
        \midrule
        PointNet\cite{pointnet} & 3.5 & 1K Points & 73.3 & 79.2 & 68.0 & 1K Points & 89.2 & - \\
        PointNet++\cite{pointnet++} & 1.5 & 1K Points & 82.3 & 84.3 & 77.9 & 1K Points & 90.7 & - \\
        DGCNN\cite{dgcnn} & 1.8 & 1K Points & 82.8 & 86.2 & 78.1 & 1K Points & 92.9 & - \\
        \midrule
        \multicolumn{9}{c}{\textit{with Self-Supervised Representation Learning (FULL)}} \\ 
        \midrule
        Point-BERT\cite{pointbert} & 22.1 & 1K Points & 87.43 & 88.12 & 83.07 & 1K Points & 92.7 & 93.2 \\
        MaskPoint\cite{Liu2022MaskPoint} & - & 2K Points & 89.30 & 88.10 & 84.30 & 1K Points & - & 93.8 \\
        Point-MAE\cite{pointmae} & 22.1 & 2K Points & 90.02 & 88.29 & 85.18 & 1K Points & 93.2 & 93.8 \\
        Point-MAE$^\dagger$\cite{pointmae} & 22.1 & 2K Points & 92.94 & 92.42 & 88.58 & 1K Points & 93.2 & 93.5 \\
        Point-M2AE\cite{point-m2ae} & 15.3 & 2K Points & 91.22 & 88.81 & 86.43 & 1K Points & 93.4 & 94.0 \\
        Point-FEMAE\cite{pointfemae} & 27.4 & 2K Points & 95.18 & 93.29 & 90.22 & 1K Points & \underline{94.0} & \textbf{94.5} \\
        PCP-MAE\cite{pcp-mae} & 22.1 & 2K Points & \underline{95.52} & \underline{93.98} & \underline{90.35} & 1K Points & \underline{94.0} & 94.2 \\
        \textbf{MPL-MAE} & 22.1 & 2K Points & \textbf{95.52} & \textbf{94.18} & \textbf{90.35} & 1K Points & \textbf{94.1} & \underline{94.4} \\
        \midrule
        \multicolumn{9}{c}{\textit{with Self-Supervised Representation Learning (MLP-LINEAR)}} \\ 
        \midrule
        Point-MAE$^\dagger$\cite{pointmae} & 22.1 & 2K Points & 86.9 & 87.3 & 76.5 & 1K Points & 90.6 & 91.2 \\
        Point-PQAE\cite{zhang2025diversechallengingpretrainingpoint} & 22.1 & 2K Points & 89.6 & \underline{90.6} & 80.9 & 1K Points & 92.2 & 92.8 \\
        Point-FEMAE\cite{pointfemae} & 27.4 & 2K Points & 89.3 & 89.2 & 80.6 & 1K Points & 92.1 & 92.0 \\
        PCP-MAE\cite{pcp-mae} & 22.1 & 2K Points & \underline{89.7} & 89.5 & \underline{81.0} & 1K Points & \underline{92.3} & \underline{93.1} \\
        \textbf{MPL-MAE} & 22.1 & 2K Points & \textbf{91.4} & \textbf{90.7} & \textbf{82.5} & 1K Points & \textbf{92.9} & \textbf{93.2} \\
        \midrule
        \multicolumn{9}{c}{\textit{with Self-Supervised Representation Learning (MLP-3)}} \\ 
        \midrule
        Point-MAE$^\dagger$\cite{pointmae} & 22.1 & 2K Points & 87.8 & 89.5 & 82.5 & 1K Points & 91.8 & 92.0 \\
        Point-PQAE\cite{zhang2025diversechallengingpretrainingpoint} & 22.1 & 2K Points & \underline{90.9} & 91.1 & 83.4 & 1K Points & 92.9 & 92.9 \\
        Point-FEMAE\cite{pointfemae} & 27.4 & 2K Points & 90.5 & \underline{91.7} & \underline{85.0} & 1K Points & 92.6 & 93.0 \\
        PCP-MAE\cite{pcp-mae} & 22.1 & 2K Points & 90.7 & 91.2 & 83.6 & 1K Points & \underline{92.9} & \underline{93.3} \\
        \textbf{MPL-MAE} & 22.1 & 2K Points & \textbf{92.9} & \textbf{92.3} & \textbf{86.0} & 1K Points & \textbf{93.0} & \textbf{93.6} \\
        \bottomrule
        \end{tabular}
    }
\end{table}

\subsection{Fine-tuning on downstream tasks}
\noindent \textbf{Object Classification.}
 We evaluate the pre-trained models under three protocols, namely MLP-Linear, MLP-3, and full fine-tuning, on ScanObjectNN and ModelNet40. Results are reported in Tab.~\ref{tab:one}.

MPL-MAE achieves competitive performance across all settings and benchmarks. Under the MLP-Linear and MLP-3 protocols, which directly reflect representation quality, MPL-MAE outperforms the previous state-of-the-art PCP-MAE by 1.2\% and 1.1\% on the OBJ-ONLY split of ScanObjectNN. On the more challenging PB-T50 split, the margins further increase to 1.5\% and 2.4\%, respectively. Although the gap narrows under full fine-tuning, MPL-MAE consistently achieves superior or comparable performance.

Compared with the baseline Point-MAE, MPL-MAE yields substantial gains, improving accuracy from 88.58\% to 90.35\% on ScanObjectNN and from 93.2\% to 94.1\% on ModelNet40. These consistent improvements demonstrate that mitigating positional leakage during pre-training leads to stronger and more transferable representations.

\begin{table}[!htpb]
    \centering
    \caption{Few-shot learning on ModelNet40. We report the average classification accuracy (\%) with the standard deviation (\%) of 10 independent experiments.}
    \label{tab:fewshot} 
    
    \resizebox{0.8\textwidth}{!}{
        \begin{tabular}{l cc cc} 
        \toprule
        \multirow{2}{*}{Methods} & \multicolumn{2}{c}{5-way} & \multicolumn{2}{c}{10-way} \\
        \cmidrule(lr){2-3} \cmidrule(lr){4-5}
        & 10-shot & 20-shot & 10-shot & 20-shot \\
        \midrule
        \multicolumn{5}{c}{\textit{Supervised Learning Only}} \\ 
        \midrule
        PointNet \cite{pointnet} & $52.0 \pm 3.8$ & $57.8 \pm 4.9$ & $46.6 \pm 4.3$ & $35.2 \pm 4.8$ \\
        DGCNN \cite{dgcnn} & $31.6 \pm 2.8$ & $40.8 \pm 4.6$ & $19.9 \pm 2.1$ & $16.9 \pm 1.5$ \\
        OcCo \cite{wang2021unsupervised} & $90.6 \pm 2.8$ & $92.5 \pm 1.9$ & $82.9 \pm 1.3$ & $86.5 \pm 2.2$ \\
        \midrule
        \multicolumn{5}{c}{\textit{with Self-Supervised Representation Learning}} \\ 
        \midrule
        Point-BERT \cite{pointbert} & $94.6 \pm 3.1$ & $96.3 \pm 2.7$ & $91.0 \pm 5.4$ & $92.7 \pm 5.1$ \\
        MaskPoint \cite{Liu2022MaskPoint} & $95.0 \pm 3.7$ & $97.2 \pm 1.7$ & $91.4 \pm 4.0$ & $93.4 \pm 3.5$ \\
        Point-MAE \cite{pointmae} & $96.3 \pm 2.5$ & $97.8 \pm 1.8$ & $92.6 \pm 4.1$ & $95.0 \pm 3.0$ \\
        Point-M2AE \cite{point-m2ae} & $96.8 \pm 1.8$ & $98.3 \pm 1.4$ & $92.3 \pm 4.5$ & $95.0 \pm 3.0$ \\
        Point-FEMAE \cite{pointfemae} & $97.2\pm 1.9$ & $98.6 \pm 1.3$ & $94.0 \pm 3.3$ & $95.8 \pm 2.8$ \\
        PCP-MAE \cite{pcp-mae} & $97.4\pm 2.3$ & $99.1 \pm 0.8$ & $93.5 \pm 3.7$ & $95.9 \pm 2.7$ \\

        \textbf{MPL-MAE} & \bm{$97.7 \pm 1.9$} & \bm{$99.2 \pm 0.9$} & \bm{$94.4 \pm 3.6$} & \bm{$96.3 \pm 2.4$} \\
        \bottomrule
        \end{tabular}
    }
\end{table}

\noindent \textbf{Few-Shot Evaluation.}
Following prior works~\cite{pointmae, pcp-mae}, we evaluate few-shot learning on ModelNet40 under the standard “$n$-way, $m$-shot” protocol, where $n \in {5, 10}$ and $m \in {10, 20}$. For each setting, the model is trained on $n \times m$ labeled samples and evaluated on 20 unseen samples per category. Results are averaged over 10 independent trials and reported with standard deviation (Tab.~\ref{tab:fewshot}).

MPL-MAE achieves state-of-the-art performance across all settings. It consistently outperforms Point-MAE by margins of 1.4\%, 1.4\%, 1.8\%, and 1.3\% across the four configurations, respectively. Moreover, compared with PCP-MAE, MPL-MAE attains higher mean accuracy with lower variance. These results demonstrate that mitigating positional leakage enhances representation robustness and sample efficiency in low-data regimes.

\noindent \textbf{Broader Evaluation.} To further validate the effectiveness and generality of MPL-MAE, we evaluate the pre-trained encoder on point cloud registration, semantic segmentation, and shape reconstruction. For registration, we follow DCP-v1~\cite{dcp} on ModelNet40 with random SE(3) transformations, replacing the DGCNN embedding network with our pre-trained encoder while keeping the remaining modules unchanged. As shown in Tab.~\ref{tab:registration}, MPL-MAE consistently outperforms Point-MAE~\cite{pointmae} and PCP-MAE~\cite{pcp-mae} across all six registration metrics, achieving $42.5\%$ and $40.0\%$ reductions in $\text{MSE}(R)$ and $\text{MSE}(t)$ over Point-MAE. We further evaluate MPL-MAE on semantic segmentation using S3DIS~\cite{s3dis} and shape reconstruction using PCN, with results reported in Tab.~\ref{tab:seg_rec}. MPL-MAE improves mAcc by $0.7\%$ over PCP-MAE and by $1.8\%$ over the baseline, while achieving substantial gains on reconstruction. These results demonstrate the effectiveness of MPL-MAE across diverse geometry-sensitive tasks.

\begin{table}[tpb]
    \centering
    \caption{Reconstruction error under context removal after pre-training.}
    \label{tab:registration} 
    
    \resizebox{\textwidth}{!}{
        \begin{tabular}{lcccccc}
        \toprule
        Method & $\text{MSE}(R)$ & $\text{RMSE}(R)$ & $\text{MAE}(R)$ & $\text{MSE}(t)$ & $\text{RMSE}(t)$ & $\text{MAE}(t)$\\

        \midrule
        Point-MAE\cite{pointmae} & 0.583 
        & 0.763 &0.494 &$2.5{\times}10^{-5}$ &$5.0{\times}10^{-3}$ &$3.7{\times}10^{-3}$ \\
        PCP-MAE \cite{pcp-mae} & 0.431 
        & 0.656 &0.410 &$1.8{\times}10^{-5}$ &$4.3{\times}10^{-3}$ &$3.1{\times}10^{-3}$ \\
        MPL-MAE & \textbf{0.335} & \textbf{0.578} &\textbf{0.366} &\boldmath{$1.5{\times}10^{-5}$} &\boldmath{$3.9{\times}10^{-3}$} &\boldmath{$2.9{\times}10^{-3}$} \\
        \bottomrule
        \end{tabular}
    }
\end{table}
\begin{table}[tpb]
    \centering
 \begin{minipage}[t]{0.35\textwidth}
        \centering
        \caption{Segmentation result and Reconstruct result.}
        \label{tab:seg_rec}
        \resizebox{\textwidth}{!}{
        \begin{tabular}{lccc}
        \toprule
        \multirow{2}{*}{Method} 
        & \multicolumn{2}{c}{Seg.} 
        & Rec. \\
        \cmidrule(lr){2-3}  \cmidrule(lr){4-4}
        &  mAcc $\uparrow$ & mIoU $\uparrow$ 
        & CD $\downarrow$ \\
        \midrule
        Point-MAE & 69.9 & 60.8 & 8.77 \\
        PCP-MAE   & 71.0 & \textbf{61.3} & 8.05 \\
        MPL-MAE   & \textbf{71.7} & 61.1 & \textbf{7.52} \\
        \bottomrule
        \end{tabular}
        }
    \end{minipage}
    \hfill 
    \begin{minipage}[t]{0.55\textwidth}
    \centering
    \captionof{table}{Effects of the main components in MPL-MAE on ScanObjectNN.}
    \label{tab:ablation1}
    \resizebox{\textwidth}{!}{
        \begin{tabular}{c c c c c c}
        \toprule
        RPE & GPI & OBJ-BG & OBJ-ONLY & PB-T50 & $RRS_{pe}$\\
        \midrule
        \ding{55} & \ding{55} & 92.94 & 92.42 & 88.58 & 0.87\\
        \ding{51} & \ding{55} & 94.49 & 93.11 & 89.41 & 0.56\\
        \ding{55} & \ding{51} & 94.32 & 92.59 & 89.31 & 0.81\\
        \ding{51} & \ding{51} & \textbf{95.52} & \textbf{94.18} & \textbf{90.35} & \textbf{0.54}\\
        \bottomrule
        \end{tabular}
    }
\end{minipage}
\end{table}

\subsection{Ablation Studies and Discussion}

\noindent \textbf{Relative Reliance Score.}
 To quantify the relative contribution of positional embeddings and encoder features to the reconstruction objective, we introduce the Relative Reliance Score (RRS). Rather than evaluating sensitivity at a single perturbation level, we measure the cumulative degradation in reconstruction performance under progressively increasing noise, thereby capturing the global influence of each component.

Formally, let $D(\sigma_x)$ denote the reconstruction loss when Gaussian noise with standard deviation $\sigma_x$ is applied to component $x \in \{pe, z\}$, while the other component remains fixed. We define
 \begin{equation}
 \mathrm{Area}(x) =
 \int_{0}^{\sigma_{\max}}
 \big[ D(\sigma_x) - D(0) \big]d\sigma_x,
 \end{equation}
 and compute the Relative Reliance Score as
 \begin{equation}
 RRS_{pe} =
 \frac{\mathrm{Area}(pe)}
 {\mathrm{Area}(pe) + \mathrm{Area}(z)}.
 \end{equation}

\noindent Here, $RRS_{pe} \in [0,1]$ measures the normalized sensitivity of reconstruction to positional perturbations. A higher value indicates stronger reliance on positional information during pre-training. We report $RRS_{pe}$ in subsequent ablations to demonstrate how different components of our framework reduce positional reliance and promote more balanced feature learning.

\noindent \textbf{Major Components.}
We conduct detailed ablation studies on RPE and GPI across multiple splits of ScanObjectNN, and report the results in Table~\ref{tab:ablation1}. 

As shown in Table~\ref{tab:ablation1}, the vanilla baseline yields a high RRS score of 0.87, indicating that reconstruction largely relies on positional embeddings. When RPE is introduced, the RRS significantly decreases to 0.56, suggesting that RPE effectively suppresses metric leakage while preserving essential topological structure. Correspondingly, downstream performance improves by 0.83\% on PB-T50 and 1.55\% on OBJ-BG. When GPI is introduced alone, the RRS decreases moderately from 0.87 to 0.81. Since GPI regulates positional injection at the decoder stage without modifying the positional encoding itself, it only partially alleviates positional dominance. Nevertheless, performance still improves by 0.73\% on PB-T50 and 1.48\% on OBJ-BG. This result indicates that GPI redistributes reconstruction supervision toward encoder features, thereby encouraging stronger semantic representation learning. When RPE and GPI are combined, the model achieves the lowest RRS and the best downstream performance across all settings. This demonstrates that structural recalibration and dynamic positional regulation are complementary, jointly mitigating positional leakage and promoting balanced optimization.
Besides, we ablate $\mathcal{L}_{\mathrm{topo}}$, whose removal reduces performance from 95.52/94.18/90.35 to 94.84/93.28/89.31 on OBJ-BG, OBJ-ONLY, and PB-T50, showing that it regularizes RPE by preserving local topology while suppressing metric-coordinate cues.

\begin{table}[!tpb]
    \centering

    \begin{minipage}[b]{0.53\textwidth}
    \centering
    \caption{Ablation study on data augmentation on ModelNet40.}
    \label{tab:data_aug_ablation}
    \resizebox{\textwidth}{!}{
        \begin{tabular}{lcccc}
        \toprule
        \multirow{2}{*}{Setting} 
        & \multicolumn{2}{c}{w/o Aug} 
        & \multicolumn{2}{c}{w/ Aug}  \\
        \cmidrule(lr){2-3}  \cmidrule(lr){4-5}
        &  Point-MAE  & MPL-MAE 
        & Point-MAE  & MPL-MAE\\
        \midrule
        FT & 92.7 & 93.7\textcolor{darkgreen}{(+1.0)} & 93.2 & 94.1\textcolor{darkgreen}{(+0.9)}\\
        MLP & 91.3 & 92.5\textcolor{darkgreen}{(+1.2)} & 91.8 & 93.0\textcolor{darkgreen}{(+1.2)}\\
        Linear & 89.5 & 92.2\textcolor{darkgreen}{(+2.7)} & 90.6 & 92.9\textcolor{darkgreen}{(+2.3)}\\
        
        \bottomrule
        \end{tabular}
    }

    \label{tab:data_aug_ablation}
\end{minipage}
\hfill
\begin{minipage}[b]{0.43\textwidth}
    \centering
    \captionof{table}{Comparison of model robustness on ModelNet40.}
    \resizebox{\textwidth}{!}{
       \begin{tabular}{lccc}
\toprule
$\sigma$ & Point-MAE & PCP-MAE & MPL-MAE \\
\midrule
0   & 93.2 & 94.0 & \textbf{94.1}$^{\textcolor{green}{+0.9}}$ \\
1.0 & 91.7 & 91.7 & \textbf{92.2}$^{\textcolor{green}{+0.5}}$ \\
2.0 & 86.7 & 87.0 & \textbf{88.3}$^{\textcolor{green}{+1.6}}$ \\
3.0 & 75.6 & 75.8 & \textbf{80.8}$^{\textcolor{green}{+5.2}}$ \\
\bottomrule
\end{tabular}
    }
    
    \label{tab:robustness}
\end{minipage}

    \begin{minipage}[t]{0.45\textwidth}
        \centering
        \caption{Ablation study on RPE candidates on ScanObjectNN.}
        \label{tab:5}
        \begin{tabular}{ll c c c}
            \toprule
            RPE &  OBJ-BG & OBJ-ONLY & PB-T50\\
            \midrule
            Euclidean&93.63 &92.94 &89.27 \\
            Cosine &\textbf{95.52} &\textbf{94.18} &\textbf{90.35} \\
            \bottomrule
        \end{tabular}
    \end{minipage}
    \hfill 
    \begin{minipage}[t]{0.45\textwidth}
        \centering
        \caption{Ablation study on GPI candidates on ScanObjectNN.}
        \label{tab:6}
        \begin{tabular}{ll c c c}
            \toprule
            GPI &  OBJ-BG & OBJ-ONLY & PB-T50 \\
            \midrule
             Sigmoid & 95.00 &93.11 &89.03 \\
            Gumbel &\textbf{ 95.52} &\textbf{94.18} &\textbf{90.35} \\
            \bottomrule
        \end{tabular}
    \end{minipage}
\end{table}

\noindent \textbf{Data Augmentation.} We further investigate the effect of data augmentation on our method across multiple settings on ModelNet40. As shown in Tab.~\ref{tab:data_aug_ablation}, when standard data augmentation is applied during pre-training, MPL-MAE improves over the baseline by 0.9\%, 1.2\%, and 2.3\% on full fine-tuning, MLP, and linear probing, respectively. When pre-training is conducted without any augmentation, the improvements remain nearly identical---1.0\%, 1.2\%, and 2.7\%. This negligible gap indicates that the gains of MPL-MAE stem from mitigating positional leakage rather than from improved robustness to data augmentation.

\noindent \textbf{Robustness.}
To further evaluate the robustness of the proposed method, we conduct noise robustness experiments on ModelNet40, comparing MPL-MAE with Point-MAE and the state-of-the-art PCP-MAE. The classification accuracy under varying levels of Gaussian noise injected into the input point clouds is reported in Tab.~\ref{tab:robustness}. By gradually increasing the noise intensity, we examine how well the learned representations remain stable under geometric perturbations.

As shown in Tab.~\ref{tab:robustness}, PCP-MAE exhibits degradation trends similar to those of Point-MAE as the noise intensity increases, indicating that enhancing supervision alone does not fundamentally alleviate positional reliance or improve robustness to geometric perturbations. In contrast, MPL-MAE consistently demonstrates the strongest robustness across all noise levels. Moreover, the performance gap widens as the noise becomes more severe, suggesting that mitigating positional leakage leads to more stable and semantically grounded representations. These results confirm that rebalancing positional and semantic contributions during pre-training enhances resilience to input perturbations and improves the stability of the learned feature space under noisy conditions.

\noindent \textbf{Candidate Operations.}
We further compare alternative design choices for RPE and GPI in Table~\ref{tab:5} and Table~\ref{tab:6} to analyze the impact of different similarity metrics and gating strategies.

From Table~\ref{tab:5}, cosine similarity consistently outperforms Euclidean distance across all ScanObjectNN variants. The RPE module aims to recalibrate supervision allocation by measuring semantic alignment between features. Unlike Euclidean distance, which is sensitive to absolute magnitudes, cosine similarity focuses on the directional alignment of embedding vectors. This property makes it more suitable for capturing structural consistency between encoder outputs and positional embeddings, thereby enabling more effective redistribution of optimization gradients.

For the gating mechanism, Table~\ref{tab:6} compares Sigmoid gating with Gumbel-Softmax. Gumbel-Softmax consistently achieves superior performance across all ScanObjectNN variants. While a standard Sigmoid provides smooth and continuous modulation, Gumbel-Softmax introduces a differentiable approximation to discrete selection. This stronger selection pressure encourages the decoder to prioritize informative encoder representations over redundant positional cues during reconstruction. As a result, it forms a more effective information bottleneck that promotes discriminative geometric feature learning, consistent with our objective of mitigating positional shortcut learning.

\section{Conclusion} 
In this paper, we proposed MPL-MAE, a masked point learning framework that mitigates positional leakage in 3D masked autoencoders. We identified an optimization imbalance that causes conventional MAE training to over-rely on coordinate cues, leading to positional shortcut learning. To address this issue, we introduced Recalibrated Positional Embedding (RPE) to suppress coordinate-specific signals while preserving geometric topology, and Gated Positional Interface (GPI) to dynamically regulate positional injection during decoding. Experiments across multiple benchmarks show that MPL-MAE achieves competitive or superior downstream performance, improved robustness to perturbations, and more balanced feature learning.

Future work will extend positional rebalancing to multimodal and large-scale geometric pre-training, and further develop a principled understanding of positional leakage under structured reconstruction objectives.

\subsubsection{Acknowledgments}
This work is partly supported by the National Key Research and Development Plan (2024YFB3309302), the National Natural Science Foundation of China (82441024), the Beijing Natural Science Foundation (L251073), the Research Program of State Key Laboratory of Complex and Critical Software Environment, and the Fundamental Research Funds for the Central Universities.

%
%
\bibliographystyle{splncs04}

\begin{thebibliography}{10}
\providecommand{\url}[1]{\texttt{#1}}
\providecommand{\urlprefix}{URL }
\providecommand{\doi}[1]{https://doi.org/#1}

\bibitem{afham2022crosspoint}
Afham, M., Dissanayake, I., Dissanayake, D., Dharmasiri, A., Thilakarathna, K., Rodrigo, R.: Crosspoint: Self-supervised cross-modal contrastive learning for 3d point cloud understanding. In: Proceedings of the IEEE/CVF conference on computer vision and pattern recognition. pp. 9902--9912 (2022)

\bibitem{s3dis}
Armeni, I., Sener, O., Zamir, A.R., Jiang, H., Brilakis, I.K., Fischer, M., Savarese, S.: 3d semantic parsing of large-scale indoor spaces. In: IEEE/CVF Conference on Computer Vision and Pattern Recognition. pp. 1534--1543 (2016)

\bibitem{chang2015shapenetinformationrich3dmodel}
Chang, A.X., Funkhouser, T., Guibas, L., Hanrahan, P., Huang, Q., Li, Z., Savarese, S., Savva, M., Song, S., Su, H., Xiao, J., Yi, L., Yu, F.: Shapenet: An information-rich 3d model repository. arXiv preprint arXiv:1512.03012  (2015)

\bibitem{chen2023pimae}
Chen, A., Zhang, K., Zhang, R., Wang, Z., Lu, Y., Guo, Y., Zhang, S.: Pi{MAE}: Point cloud and image interactive masked autoencoders for {3D} object detection. In: Proceedings of the IEEE/CVF Conference on Computer Vision and Pattern Recognition. pp. 5291--5301 (2023)

\bibitem{chen2023pointgpt}
Chen, G., Wang, M., Yang, Y., Yu, K., Yuan, L., Yue, Y.: Pointgpt: Auto-regressively generative pre-training from point clouds. In: Advances in Neural Information Processing Systems (2023)

\bibitem{chhipa2022depthcontrastselfsupervisedpretraining}
Chhipa, P.C., Upadhyay, R., Saini, R., Lindqvist, L., Nordenskjold, R., Uchida, S., Liwicki, M.: Depth contrast: {Self-supervised} pretraining on {3DPM} images for mining material classification. In: Proceedings of the European Conference on Computer Vision Workshops. pp. 212--227 (2022)

\bibitem{dong2023autoencoderscrossmodalteacherspretrained}
Dong, R., Qi, Z., Zhang, L., Zhang, J., Sun, J., Ge, Z., Yi, L., Ma, K.: Autoencoders as {Cross-Modal} teachers: Can pretrained {2D} image transformers help {3D} representation learning? In: Proceedings of the International Conference on Learning Representations (2023)

\bibitem{fan2017point}
Fan, H., Su, H., Guibas, L.: A point set generation network for 3d object reconstruction from a single image. In: Proceedings of the IEEE/CVF Conference on Computer Vision and Pattern Recognition. pp. 2463--2471 (2017)

\bibitem{byol}
Grill, J.B., Strub, F., Altché, F., Tallec, C., Richemond, P.H., Buchatskaya, E., Doersch, C., Pires, B.A., Guo, Z.D., Azar, M.G., Piot, B., Kavukcuoglu, K., Munos, R., Valko, M.: Bootstrap your own latent a new approach to self-supervised learning. In: Advances in Neural Information Processing Systems (2020)

\bibitem{guo2023joint}
Guo, Z., Zhang, R., Qiu, L., Li, X., Heng, P.A.: Joint-mae: 2d-3d joint masked autoencoders for 3d point cloud pre-training. In: Proceedings of the International Joint Conference on Artificial Intelligence (2023)

\bibitem{csc}
Hou, J., Graham, B., Nie{\ss}ner, M., Xie, S.: Exploring data-efficient 3d scene understanding with contrastive scene contexts. In: Proceedings of the IEEE/CVF conference on computer vision and pattern recognition. pp. 15587--15597 (2021)

\bibitem{huang2021spatio}
Huang, S., Xie, Y., Zhu, S.C., Zhu, Y.: Spatio-temporal self-supervised representation learning for 3d point clouds. In: Proceedings of the IEEE/CVF international conference on computer vision. pp. 6535--6545 (2021)

\bibitem{li2022semmae}
Li, G., Zheng, H., Liu, D., Wang, C., Su, B., Zheng, C.: Semmae: Semantic-guided masking for learning masked autoencoders. In: Advances in Neural Information Processing Systems (2022)

\bibitem{Liu2022MaskPoint}
Liu, H., Cai, M., Lee, Y.J.: Masked discrimination for self-supervised learning on point clouds. In: Proceedings of the European Conference on Computer Vision. pp. 657--675 (2022)

\bibitem{liu2024inter}
Liu, J., Wu, Y., Gong, M., Liu, Z., Miao, Q., Ma, W.: Inter-modal masked autoencoder for self-supervised learning on point clouds. IEEE Transactions on Multimedia  \textbf{26},  3897--3908 (2024)

\bibitem{liu2026dcmae}
Liu, X., Wang, F., Chen, Z., Dong, X.: Dcmae: A dual-branch contrastive masked autoencoder for 3d object detection. Journal of Visual Communication and Image Representation  \textbf{115},  104675 (2026)

\bibitem{loshchilov2017decoupled}
Loshchilov, I., Hutter, F.: Decoupled weight decay regularization. arXiv preprint arXiv:1711.05101  (2017)

\bibitem{loshchilov2017sgdrstochasticgradientdescent}
Loshchilov, I., Hutter, F.: Sgdr: Stochastic gradient descent with warm restarts. arXiv preprint arXiv:1608.03983  (2017)

\bibitem{luo2021self}
Luo, C., Yang, X., Yuille, A.: Self-supervised pillar motion learning for autonomous driving. In: Proceedings of the IEEE/CVF Conference on Computer Vision and Pattern Recognition. pp. 3183--3192 (2021)

\bibitem{mersch2022self}
Mersch, B., Chen, X., Behley, J., Stachniss, C.: Self-supervised point cloud prediction using 3d spatio-temporal convolutional networks. In: Proceedings of the Conference on Robot Learning (2021)

\bibitem{min2023occupancy}
Min, C., Xu, X., Zhao, D., Xiao, L., Nie, Y., Dai, B.: Occupancy-mae: Self-supervised pre-training large-scale lidar point clouds with masked occupancy autoencoders. arXiv preprint arXiv:2206.09900  (2023)

\bibitem{pointmae}
Pang, Y., Wang, W., Tay, F.E.H., Liu, W., Tian, Y., Yuan, L.: Masked autoencoders for point cloud self-supervised learning. In: Proceedings of the European Conference on Computer Vision. pp. 604--621 (2022)

\bibitem{peng2022self}
Peng, G., Ren, Z., Wang, H., Li, X., Khyam, M.O.: A self-supervised learning-based 6-dof grasp planning method for manipulator. IEEE Transactions on Automation Science and Engineering  \textbf{19}(4),  3639--3648 (2022)

\bibitem{dgcnn}
Phan, A.V., Le~Nguyen, M., Nguyen, Y.L.H., Bui, L.T.: Dgcnn: A convolutional neural network over large-scale labeled graphs. Neural Networks  \textbf{108},  533--543 (2018)

\bibitem{pointnet}
Qi, C.R., Su, H., Mo, K., Guibas, L.J.: Pointnet: Deep learning on point sets for 3d classification and segmentation. In: Proceedings of the IEEE conference on computer vision and pattern recognition. pp. 652--660 (2017)

\bibitem{pointnet++}
Qi, C.R., Yi, L., Su, H., Guibas, L.J.: Pointnet++: Deep hierarchical feature learning on point sets in a metric space. Advances in neural information processing systems  \textbf{30} (2017)

\bibitem{qi2023contrastreconstructcontrastive3d}
Qi, Z., Dong, R., Fan, G., Ge, Z., Zhang, X., Ma, K., Yi, L.: Contrast with reconstruct: Contrastive 3d representation learning guided by generative pretraining. In: Proceedings of the International Conference on Machine Learning (2023)

\bibitem{sautier2022image}
Sautier, C., Puy, G., Gidaris, S., Boulch, A., Bursuc, A., Marlet, R.: Image-to-lidar self-supervised distillation for autonomous driving data. In: Proceedings of the IEEE/CVF Conference on Computer Vision and Pattern Recognition. pp. 9891--9901 (2022)

\bibitem{scanobjectNN}
Uy, M.A., Pham, Q.H., Hua, B.S., Nguyen, T., Yeung, S.K.: Revisiting point cloud classification: A new benchmark dataset and classification model on real-world data. In: Proceedings of the IEEE/CVF International Conference on Computer Vision. pp. 1588--1597 (2019)

\bibitem{wang2021unsupervised}
Wang, H., Liu, Q., Yue, X., Lasenby, J., Kusner, M.J.: Unsupervised point cloud pre-training via occlusion completion. In: Proceedings of the IEEE/CVF international conference on computer vision. pp. 9782--9792 (2021)

\bibitem{mrd}
Wang, H., Bao, Y., Pan, P., Li, Z., Liu, X., Yang, R., Huang, D.: Multi-modal relation distillation for unified {3D} representation learning. In: Proceedings of the European Conference on Computer Vision. pp. 364--381 (2024)

\bibitem{dcp}
Wang, Y., Solomon, J.M.: Deep closest point: Learning representations for point cloud registration. In: Proceedings of the IEEE/CVF international conference on computer vision. pp. 3522--3531 (2019)

\bibitem{wu2020pointpwc}
Wu, W., Wang, Z.Y., Li, Z., Liu, W., Fuxin, L.: Pointpwc-net: Cost volume on point clouds for (self-)supervised scene flow estimation. In: Proceedings of the European Conference on Computer Vision. pp. 88--107 (2020)

\bibitem{wu2023masked}
Wu, X., Wen, X., Liu, X., Zhao, H.: Masked scene contrast: A scalable framework for unsupervised 3d representation learning. In: Proceedings of the IEEE/CVF Conference on computer vision and pattern recognition. pp. 9415--9424 (2023)

\bibitem{ModelNet40}
Wu, Z., Song, S., Khosla, A., Yu, F., Zhang, L., Tang, X., Xiao, J.: 3d shapenets: A deep representation for volumetric shapes. In: Proceedings of the IEEE Conference on Computer Vision and Pattern Recognition. pp. 1912--1920 (2015)

\bibitem{pointcontrast}
Xie, S., Gu, J., Guo, D., Qi, C.R., Guibas, L., Litany, O.: Pointcontrast: Unsupervised pre-training for 3d point cloud understanding. In: European conference on computer vision. pp. 574--591 (2020)

\bibitem{ulip}
Xue, L., Gao, M., Xing, C., Mart{\'\i}n-Mart{\'\i}n, R., Wu, J., Xiong, C., Xu, R., Niebles, J.C., Savarese, S.: Ulip: Learning a unified representation of language, images, and point clouds for 3d understanding. In: Proceedings of the IEEE/CVF conference on computer vision and pattern recognition. pp. 1179--1189 (2023)

\bibitem{ulip2}
Xue, L., Yu, N., Zhang, S., Panagopoulou, A., Li, J., Mart{\'\i}n-Mart{\'\i}n, R., Wu, J., Xiong, C., Xu, R., Niebles, J.C., et~al.: Ulip-2: Towards scalable multimodal pre-training for 3d understanding. In: Proceedings of the IEEE/CVF Conference on Computer Vision and Pattern Recognition. pp. 27091--27101 (2024)

\bibitem{yu2024robotic}
Yu, S., Zhai, D.H., Xia, Y.: Robotic grasp detection based on category-level object pose estimation with {Self-Supervised} learning. IEEE/ASME Transactions on Mechatronics  \textbf{29}(1),  625--635 (2023)

\bibitem{pointbert}
Yu, X., Tang, L., Rao, Y., Huang, T., Zhou, J., Lu, J.: Point-bert: Pre-training 3d point cloud transformers with masked point modeling. In: IEEE/CVF Conference on Computer Vision and Pattern Recognition. pp. 19291--19300 (2022)

\bibitem{pointfemae}
Zha, Y., Ji, H., Li, J., Li, R., Dai, T., Chen, B., Wang, Z., Xia, S.T.: Towards compact 3d representations via point feature enhancement masked autoencoders. In: Proceedings of the AAAI conference on artificial intelligence. vol.~38, pp. 6962--6970 (2024)

\bibitem{point-m2ae}
Zhang, R., Guo, Z., Fang, R., Zhao, B., Wang, D., Qiao, Y., Li, H., Gao, P.: Point-m2ae: Multi-scale masked autoencoders for hierarchical point cloud pre-training. In: Advances in Neural Information Processing Systems. vol.~35, pp. 27061--27074 (2022)

\bibitem{zhang2023learning}
Zhang, R., Wang, L., Qiao, Y., Gao, P., Li, H.: Learning 3d representations from 2d pre-trained models via image-to-point masked autoencoders. In: Proceedings of the IEEE/CVF conference on computer vision and pattern recognition. pp. 21769--21780 (2023)

\bibitem{pcp-mae}
Zhang, X., Zhang, S., Yan, J.: Pcp-mae: Learning to predict centers for point masked autoencoders. In: Advances in Neural Information Processing Systems. vol.~37, pp. 80303--80327 (2024)

\bibitem{zhang2025diversechallengingpretrainingpoint}
Zhang, X., Zhang, S., Yan, J.: Towards more diverse and challenging pre-training for point cloud learning: Self-supervised cross reconstruction with decoupled views. In: Proceedings of the IEEE/CVF International Conference on Computer Vision (2025)

\bibitem{uni3d}
Zhou, J., Wang, J., Ma, B., Liu, Y.S., Huang, T., Wang, X.: Uni3d: Exploring unified 3d representation at scale. In: Proceedings of the International Conference on Learning Representations (2023)

\end{thebibliography}

\end{document}